# New hard benchmark functions for global optimization

Abdesslem Layeb
Laboratory of data science and artificial intelligence (LISIA)
university of Constantine 2, department of fundamental computer science and its applications
Abdesslem.layeb@univ-constantine2.dz

**Abstract.** In this paper, we present some new unimodal, multimodal, and noise test functions to assess the performance of global optimization algorithms. All the test functions are multidimensional problems. The 2-dimension landscape of the proposed functions has been graphically presented in 3D space to show their geometry, however these functions are more complicated in dimensions greater than 3. To show the hardness of these functions, we have made an experimental study with some powerful algorithms such as CEC competition winners: LSHADE, MadDe, and LSHADE-SPACMA algorithms. Besides the novel algorithm, Tangent search algorithm (TSA) and its modified Tangent search algorithm (mTSA) were also used in the experimental study. The results found demonstrate the hardness of the proposed functions.

The code sources of the proposed test functions are available on Matlab Exchange website.

**Keywords:** test function optimization, global optimization.

1. Introduction

We present in this work 20 test functions with nontrivial global minima: 2 unimodal functions, 16 multimodal functions, and 2 noisy functions. The motivations of this work are multiple.

First, the proposed unimodal functions have very large flat surfaces, and so they do not help the algorithms to direct the search process towards the minimum. We have noted that These functions are very hard for differential-evolution based algorithms.

Second, we propose hard multimodal functions with known global optima which are not necessarily sequence of 0 like in Ackley or Rastrigin functions in order to avoid favoring algorithms whose search process is oriented towards the center of the search space. Besides, we aim to show the precision of algorithms by optimizing functions whose optimal solutions are a multiple of $\pi$, fraction of $\pi$, integer values, or functions with a small power order like 0.1.

Third, and it is the most important motivation, in the literature, several test functions claimed by its authors very hard; but they are not really so hard. The problem is not due to the low performance of the algorithms but to the low accuracy of the programming language. Let's take the example of the function crosslegtable (eq.1) claimed to be hard by the author Mishra (Mishra, 2006). This function was ranked third in the hardest test functions after DeVilliersGlasser02 and Damavandi test functions according to the work of (Gavana, 2009). They say that the global optimum of this function is –1 and that the optimum solution is $x_i$=0 for 2-dimension. Nevertheless, this is not the only global optimum, it is enough that at least one variable is equal to $k\pi$ to get one of its global optima where k is an integer. On the other side, if we implement this function with Matlab programming language, the function is very hard to optimize, it is composed of complex mathematical functions like sine, exponential and square root, power of 0.1, and $\pi$ constant (eq1). However, the difficulty of this function is not due really to the hardness of the function or to the stagnation of algorithms in local minima, but to the accuracy of the programming language in computing the numerical functions such as sine/cosine, exponential, power...etc, and some mathematical constants like $\pi$. The truth is that the algorithms are "got stuck !" in one of its nonzero global optima such as [π, π] and this optimum doesn't give the optimum value –1 by using the Matlab language. Let's explain more, for example, if we compute sin(π) in Matlab, we obtain 1.224646799147353e-16 not a perfect 0, and 1.224646799147353e-16$^{0.1}$ is equal to 0.025633097248031 very far from zero, the same observation for abs(2-($2^{0.5}$)$^2$)$^{0.1}$= 2.9157e–02! computed by Matlab and not a 0. This inaccuracy of the Matlab language will give a wrong orientation to any optimizer algorithm and make some easy functions hard to optimize. Concerning crosslegtable, the value of this function at the point [π, π] is –0.079592386218981 not –1 even that [π, π] is mathematically one of the global optima of this function. What is the solution for this problem? There are two solutions, the first is to increase the precision of π by using the Variable-precision arithmetic function *vpa()* in Matlab:

*(sin(vpa(pi,32))= –3.2101083013100396069547145883568e-40)*

or using symbolic variable sym('pi'), unfortunately, this solution will decrease the speed of the algorithms and it is impracticable. The second solution that we propose is to use the degrees instead of radians. Indeed, the sin(180°)=0 more accurate than sin(π), so we should transform each variable from radian to degree and use sind function instead of the radian sin function by using the transformation in equation 2.

$$f(X) = -\frac{1}{(|\sin(x_1)\sin(x_2)e^{|100-\frac{\sqrt{x_1^2+x_2^2}}{\pi}|}|+1)^{0.1}} \quad\text{............(eq1)}$$

$$xd_i = \frac{x_i}{\pi}180 \quad\text{....................................................(eq.2)}$$

Where $x_i$ is the decision variable.

By using this transformation, we are able to optimize this function successfully. This remark is valid for several test functions in the literature. Concerning the problem of power accuracy, it is recommended to not use many imbrications of the function power greater than three imbrications.

Finally, two new noisy functions are presented to assess the robustness of the optimization algorithms in stochastic environments.

In this context, we present in this work 20 test functions designed to challenge the most powerful algorithms among of the winners in the CEC competitions such as Improving the Search Performance of SHADE Using Linear Population Size Reduction (LSHADE) (Tanabe, 2014), Improving differential evolution through bayesian hyperparameter optimization (MadDe) (Biswas,2021) and LSHADE-SPACMA algorithm: an hybridation of LSHADE and the famous Covariance matrix adaptation evolution strategy (CMAES) algorithm (Mohamed,2017). Besides, a recent optimization algorithm the Tangent search algorithm, (TSA)( Layeb, 2022) and its modified version mTSA are used in this evaluation. The experimental results show that the proposed functions represent a real challenge to optimization algorithms, especially in dimensions greater than 30.

2. **The new test functions**

In the following, we present the developed functions. The 3d surfaces of the proposed functions are displayed in figure 1.

- **Layeb01 function:** (unimodal, separable and scalable). This is a unimodal function with large search range which pose a challenge for the algorithms. the global minimum has a small area relative to the large search space.

- The optimal value is f(x*)= 0.
- The global minimum $x_i^*$ =1;
- The search domain is [-100, 100]

$$f(x) = \sum_{i}^{n} 100 \sqrt[2]{|e^{(x_i-1)^2} - 1|}$$

- **Layeb02 function:** (unimodal, separable and scalable). This is unimodal function; it is defined as follow. This function has high values around the [0,0,….], and the global minimum is located at the point [1,1,… ] which poses a problem to the optimizer algorithms.
    - The optimal value is f(x*)= 0.
    - The global minimum $x_i^*$ =1;
    - The search domain is [-10, 10]

$$f(x) = \sum_{i=1}^{n} |e^{100\frac{(x_i-1)^2}{e^{x_i+1}}} - 1|$$

- **Layeb03 function (Arclegtable):** (multimodal, non-separable and scalable). This function is a modification of crosslegtable function, and it is defined as follow. This function is very hard to optimize in higher dimension.
    - The optimal value is f(x*)= –n+1
    - The global minimum is $x_i^* = k\pi$, k is an integer
    - The search domain is [-10, 10]

$$f(x) = \sum_{i}^{n-1} |\sin(x_i) e^{\left|100 - \frac{\sqrt{x_i^2 + x_{i+1}^2}}{\pi}\right|} + \sin(x_{i+1}) + 1|^{-0.1}$$

- **Layeb04 function (Crossfly):** (multimodal, non-separable and scalable). This function is defined as follow
    - The optimal value is f(x*)= (ln(0.001)-1)(dim-1).
    - The global minimum $x_i^*$ is alternation of 0 and (2k-1)π, k is an integer.
    - The search domain is [-10, 10]

$$f(x) = \sum_{i}^{n-1} \ln(|x_i x_{i+1}| + 0.001) + \cos(x_i + x_{i+1})$$

- **Layeb05 function (Dome):** (multimodal, non-separable and scalable). This function is defined as follow.
    - The optimal value is f(x*)= ln(0.001) (n-1).
    - The global minimum is x*= $((2k-1)\pi, 2k\pi, (2k-1)\pi, 2k\pi ... ...)$ or$(2k\pi, (2k-1)\pi, 2k\pi, (2k-1)\pi, ... ...)$, k is an integer
    - The search domain is [-10, 10]

$$f(x) = \sum_{i}^{n-1} \frac{\text{lin}(|\sin\left(x_i - \frac{\pi}{2}\right) + \cos(x_{i+1} - \pi)| + 0.001)}{|\cos(2x_i - x_{i+1} + \frac{\pi}{2})| + 1}$$

- **Layeb06 function (Infinity ):** (multimodal, non-separable and scalable). This function is defined as follow.
    - The optimal value is f(x*)= 0.
    - The global minimum is $x_i^* = (2k-1)\pi$, k is an integer
    - The search domain is [-10, 10]

$$f(x) = \sum_{i}^{n-1} (|\cos(\sqrt{x_i^2 + x_{i+1}^2}) \sin(x_{i+1}) + \cos(x_{i+1}) + 1|)^{0.1}$$

- **Layeb07 function:** (multimodal, non-separable and scalable). This function is defined as follow.
    - The optimal value is f(x*)= 0.
    - The global minimum is located at $x_i^* = \frac{(2k-1)\pi}{2}$ and $x_i^* = k\pi$, k is an integer
    - The search domain is [-10, 10]

$$f(x) = \sum_{i}^{n-1} (100 |\cos(x_i + x_{i+1} - \frac{\pi}{2})|^{0.1} - e^{\cos(\frac{16 x_i x_{i+1}}{\pi})} + e^1)$$

- **Layeb08 function:** (multimodal, non-separable and scalable). This hard function is defined as follow.
    - The optimal value is f(x*)= log(0.001)(n-1).
    - The global minimum x* is an alternation of π/4,–π/4
    - The search domain is [-10, 10]

$$f(x) = \sum_{i}^{n-1} \ln(|x_i - x_{i+1}| + 0.001) + |(100 \cos(x_i - x_{i+1})|)$$

- **Layeb09 function:** This function is defined as follow.
    - The optimal value is f(x*)= 0.
    - The global minimum is $x_i^* = \frac{(2k-1)\pi}{2}$, k is an integer
    - The search domain is [-10, 10]

$$f(x) = \sum_{i}^{n-1} \sqrt[2]{\frac{e^{(|x_{i+1}|(|\sin(x_{i+1})|-1))} + \cos(x_i + x_{i+1})}{e^{\cos(x_i + x_{i+1})-1}}}$$

- **Layeb10 function:** (multimodal, non-separable and scalable). This hard function is defined as follow.
    - The optimal value is f(x*)= 0.
    - The global minimum x* =[0.5 0.5 0.5……]
    - The search domain is [-100, 100]

$$f(x) = \sum_{i}^{n-1} (((\ln(x_i^2 + x_{i+1}^2 + 0.5))^2 + |(100 \sin(x_i - x_{i+1})|)$$

- **Layeb11 function:** (multimodal, non-separable and scalable). This hard function is defined as follow.
    - The optimal value is f(x*)=–n+1.
    - The global minimum x* is an alternation of –1 and 0
    - The search domain is [-10, 10]

$$f(x) = \sum_{i}^{n-1} \frac{\cos(x_i * x_{i+1} + \pi)}{(100 * |x_i^2 - x_{i+1} - 1|)^2 + 1}$$

- **Layeb12 function:** (multimodal, non-separable and scalable). This function is defined as follow. This function is very hard to optimize because the optimizer should give an integer or near-integer equal to 2. The most algorithms fail in this test.

- The optimal value is f(x*)= – (e+1)(n-1).
- The global minimum is $x_i^*$=2.
- The search domain is [-5, 5]

$$f(x) = -\sum_{i}^{n-1}\left(\cos\left(\frac{\pi}{2}x_i - \frac{\pi}{4}x_{i+1} - \frac{\pi}{2}\right)e^{\cos(2\pi x_i x_{i+1})} + 1\right)$$

- **Layeb13 function:** (multimodal, non-separable and scalable). This function is defined as follow. This function is very hard to optimize; all the algorithms fail in this test.
  - The optimal value is f(x*)= 0.
  - The global minimum is x* is an alternation of (2k+1)π/4,–(2k+1)π/4, k is an integer.
  - The search domain is [-10, 10]

$$f(x) = \sum_{i}^{n-1}(|\cos(x_i - x_{i+1})| + 100|\log(|x_i + x_{i+1}| + 1)|^{0.1})$$

- **Layeb14 function:** (multimodal, non-separable and scalable). This is a hard function; the most algorithms fail in this test.
  - The optimal value is f(x*)= 0.
  - The global minimum is x* is a permutation of 0 and –1.
  - The search domain is [-100, 100]

$$f(x) = -\sum_{i}^{n-1}(100|x_i^2 - x_{i+1} - 1|^{0.1} + |\log((x_i + x_{i+1} + 2)^2)|)$$

- **Layeb15 function:** (multimodal, non-separable and scalable). This function is defined as follow. This function is very hard to optimize, the most algorithms fail in this test.
  - The optimal value is f(x*)= 0.
  - The global minimum is an alternation of 1 and –1.
  - The search domain is [-100, 100]

$$f(x) = \sum_{i}^{n-1}(10\sqrt{\tanh(2|x_i| - x_{i+1}^2 - 1)} + |e^{(x_i * x_{i+1}+1)} - 1|)$$

- **Layeb16 function (Marmar):** (multimodal, non-separable and scalable). This function is defined as follow:
  - The optimal value is f(x*)= 0.
  - The global minimum is $x_i^*$= π/4 or $x_i^*$= −π/4
  - The search domain is [-10, 10]

$$f(x) = -\sum_{i}^{n-1}(|\tan(x_{i+1})x_i + 100|\cos(x_i)^2 - \sin(x_{i+1})^2)| - \frac{\pi}{4}|)^{0.2}$$

- **Layeb17 function (Wings):** (multimodal, non-separable and scalable). This function is defined as follow:
  - The optimal value is f(x*)= 0.
  - The global minimum x* is an alternation of -1 and 0.
  - The search domain is [-100, 100]

$$f(x) = \sum_{i}^{n-1} \left(10 * |\ln((x_i + x_{i+1} + 2)^2| - \frac{1}{\left(1000 * (x_i^2 - x_{i+1} - 1)\right)^2 + 1} + 1\right)$$

- **Layeb18 function (Zohra):** (multimodal, non-separable and scalable). This function is defined as follow:
    - The optimal value is f(x*)= ln(0.001) (n-1).
    - The global minimum is $x_i$*= $(2k - 1)\frac{\pi}{2}$, k is an integer
    - The search domain is [-10, 10]

$$f(x) = \sum_{i}^{n-1} \frac{\ln(|\cos(2x_i x_{i+1}/\pi)| + 0.001)}{|\sin(x_i + x_{i+1})\cos(x_i)| + 1}$$

- **Layeb19 function (Noiselog):** (unimodal, separable and scalable). This function is very hard to optimize; it contains a random component. This function is defined as follow.
    - The optimal value is f(x*)= 0.
    - The global minimum is $x_i$*=1
    - The search domain is [-5, 5].

$$f(x) = \sum_{i}^{n} 100 rand^i (\ln((x_i - 1)^2 + 1))^2$$

- **Layeb20 function (Noisesphere):** (unimodal, separable and scalable). This function is very hard to optimize; it contains a random component. This function is defined as follow.
    - The optimal value is f(x*)= 0.
    - The global minimum is $x_i$*=1
    - The search domain is [-5, 5].

$$f(x) = \sum_{i}^{n} rand^i (x_i - 1)^2$$

| **Layeb01** 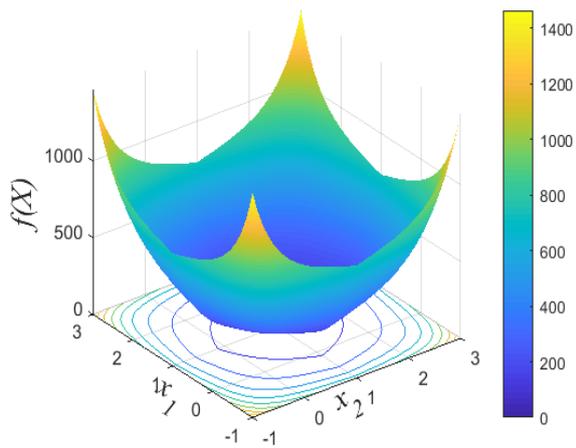 | **Layeb02** 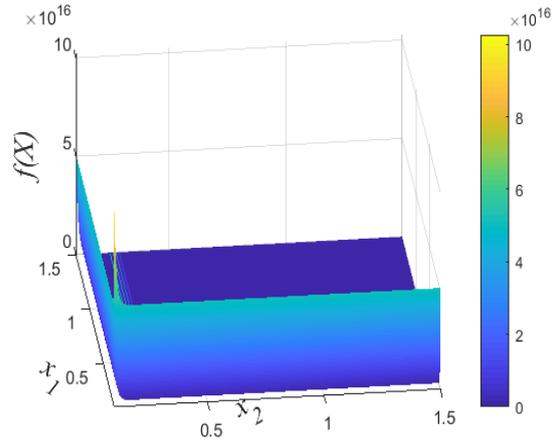 |
|---|---|
| **Layeb03** 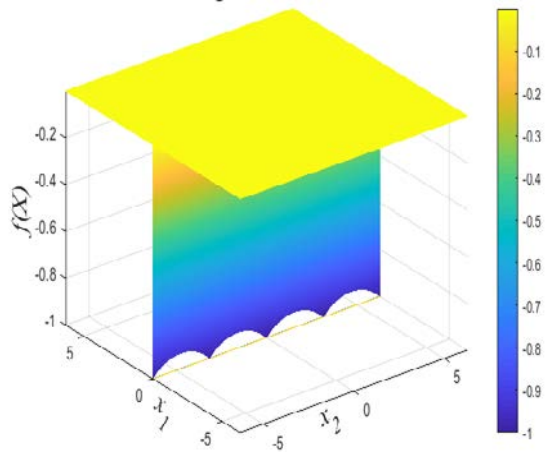 | **Layeb04** 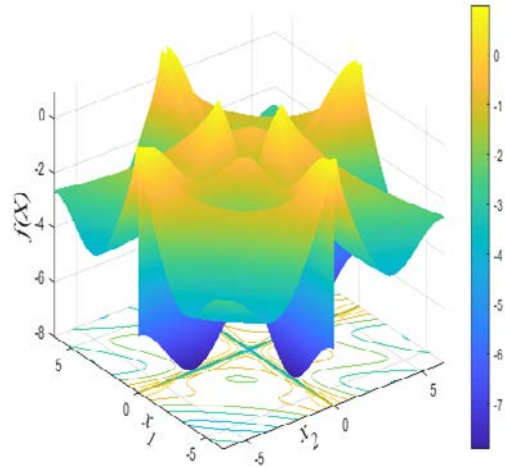 |
| **Layeb05** 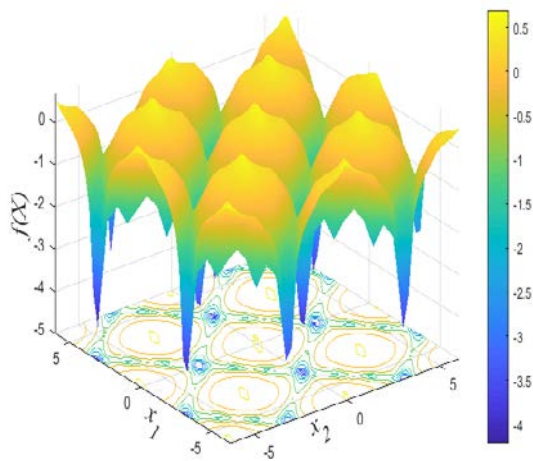 | **Layeb06** 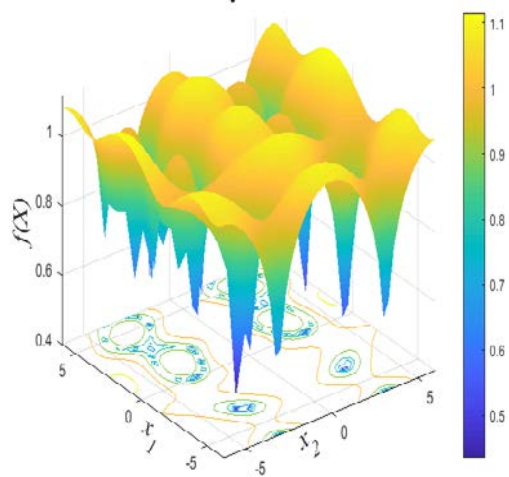 |

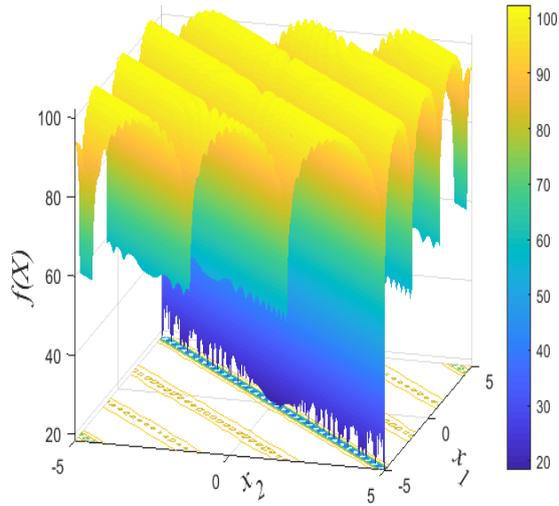
Layeb07

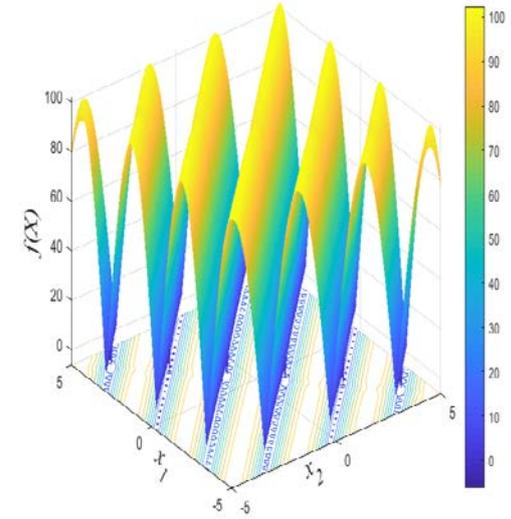
Layeb08

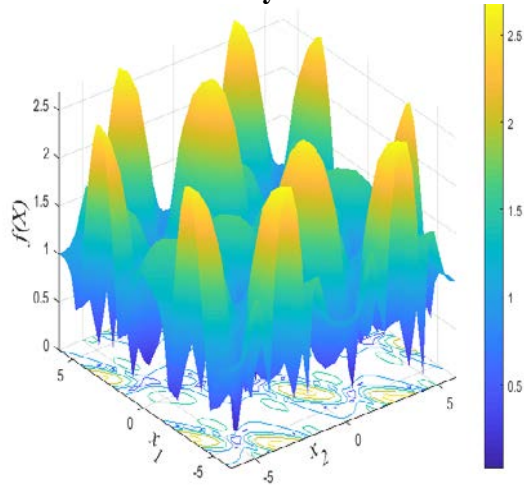
Layeb09

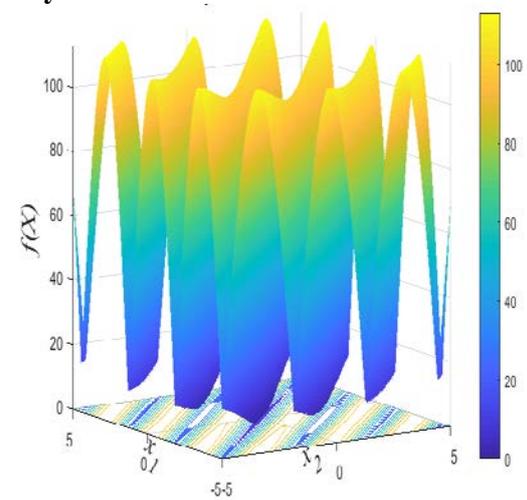
Layeb10

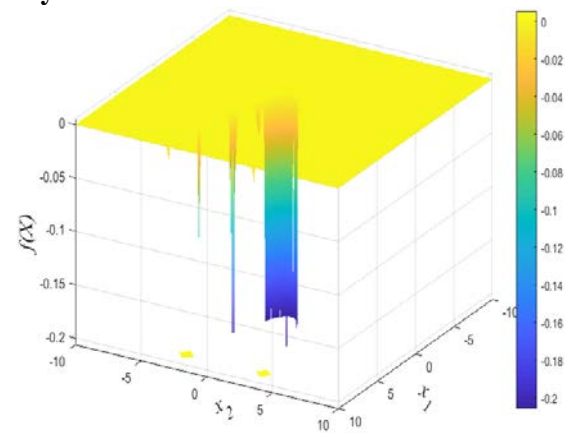
Layeb11

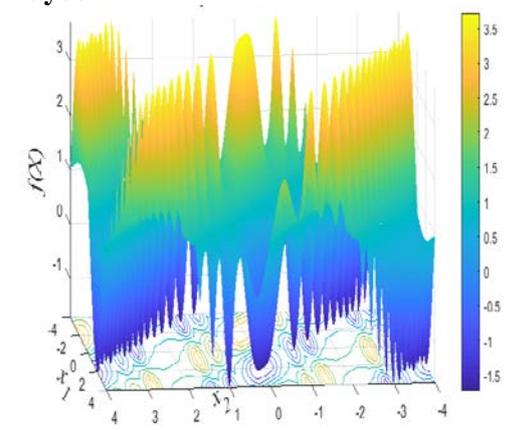
Layeb12

| **Layeb13** | **Layeb14** |
|---|---|
| 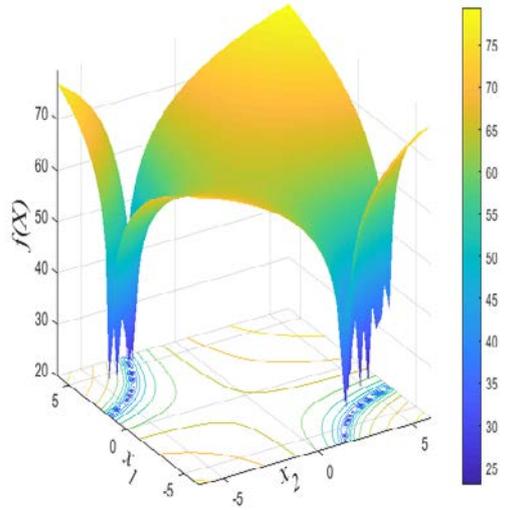 | 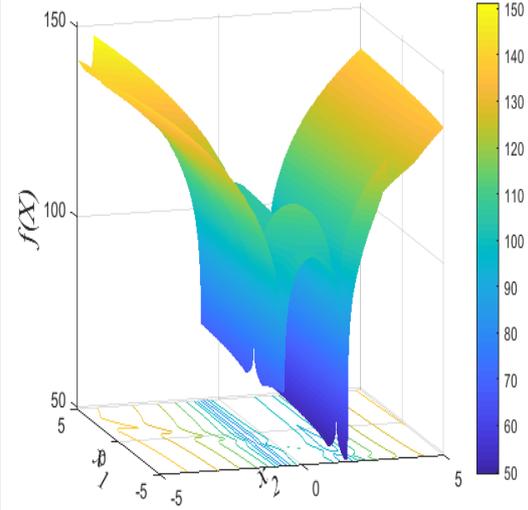 |
| **Layeb15** | **Layeb16** |
| 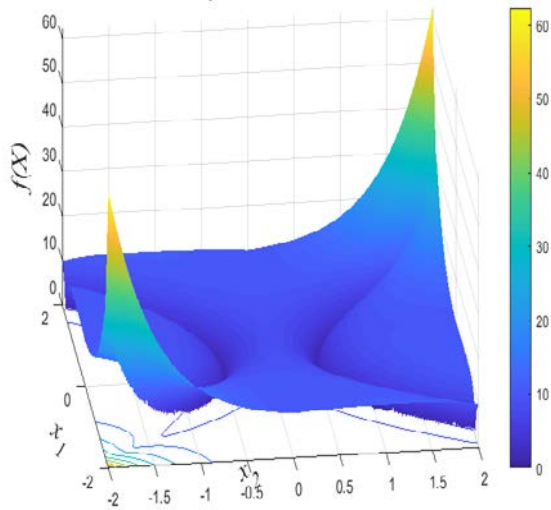 | 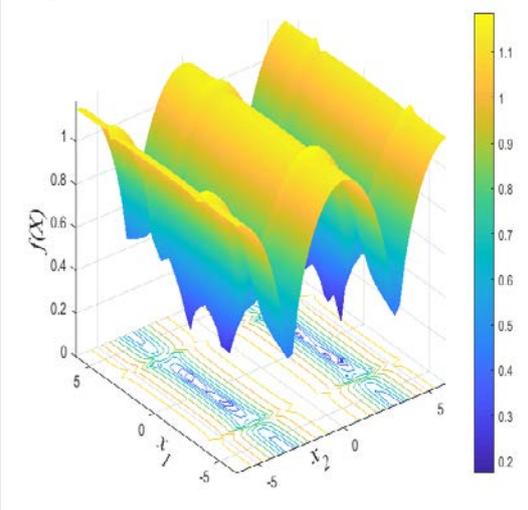 |
| **Layeb17** | **Layeb18** |
| 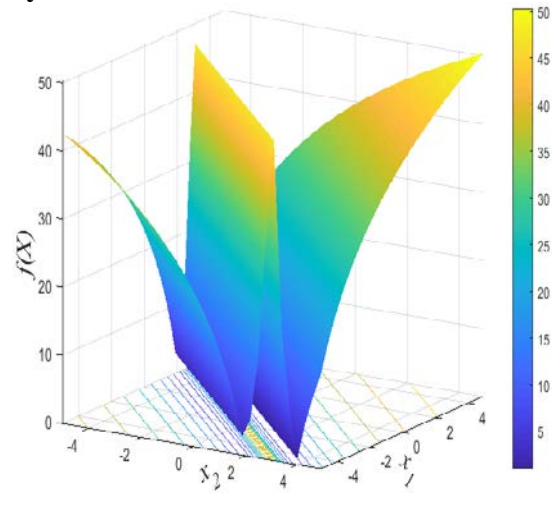 | 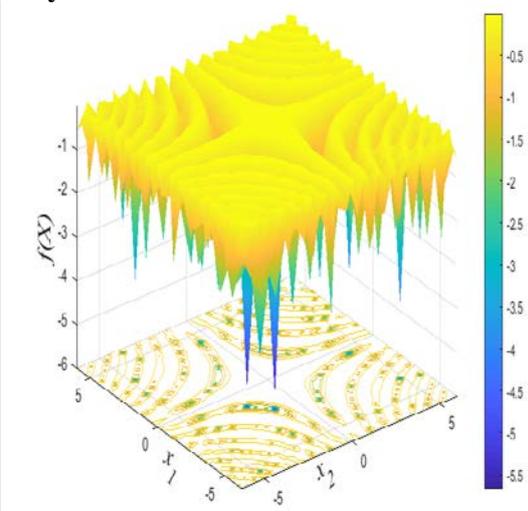 |

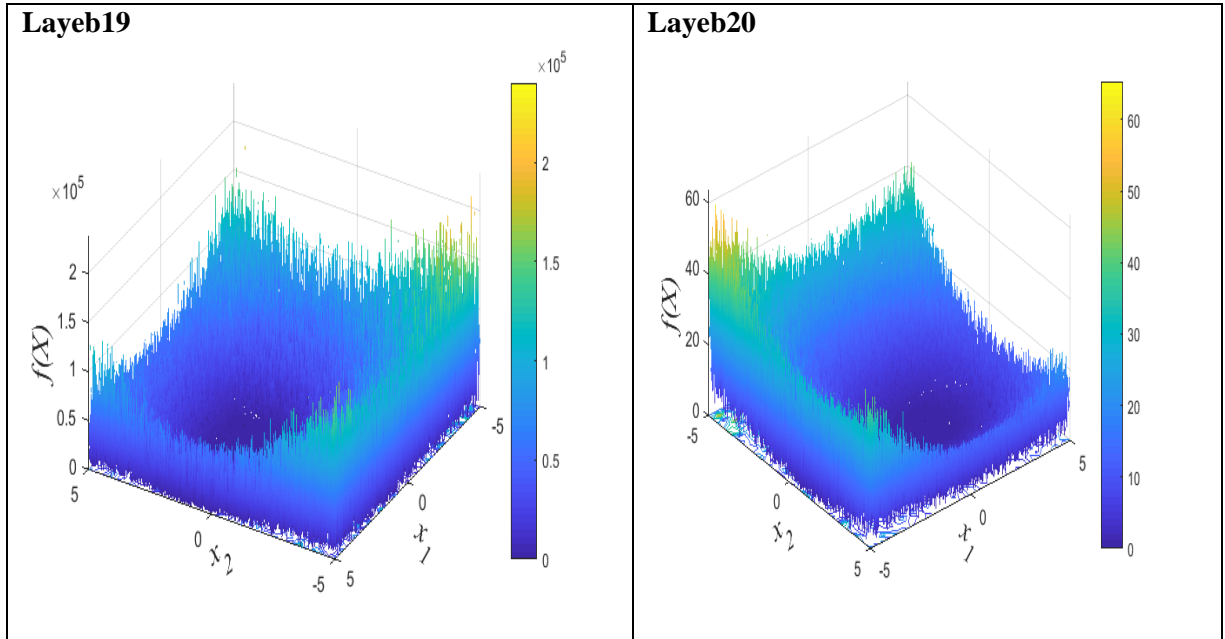

**Figure 1.** The 3d surfaces of the proposed functions

## 3. Experimental results

All the algorithms, mTSA, TSA, LSHADE , MadDe, and LSHADE-SPACMA, used in this experiment are encoded in Matlab and we have used their default parameters as in the CEC competitions.  The LSHADE , MadDe, and LSHADE-SPACMA are downloaded from the  repository  of P-N-Suganthan (https://github.com/P-N-Suganthan). For fair comparison, each algorithm is executed 30 times for 30 and 10  dimensions for each test function. The maximum number of function evaluations is set to $10^4*D$. The mean, std, min (best) and max (worst) results are computed.  Finally, Friedman tests were carried out to show the difficulty of the proposed tests and which algorithms performed well in this experiment. It should be noted that all the results were normalized before using Friedman tests in order to get a good interpretation. The results for dimensions 30d and 10d  are summarized  in tables 1 and 2.

Globally, the mean results of all the algorithms are far from the global optima. In the 30d function evaluations experiment (table 2), mTSA, TSA, and LSHADE-SPACMA are the nearest to the global optima as confirmed by the Friedman test ( figure 2). However, LSHADE and MadDe algorithms rank last in this experiment.  On the other hand, on 10d test functions (table 2), LSHADE-SPACMA ranks first, the other algorithms have similar performance as it is confirmed by Friedman test (figure 3)

Concerning the hardness of the proposed functions. As we can see, all the algorithms fail to find the exact optimum in most test functions. In the unimodal functions ( Layeb01 and Layeb02) the  DE based algorithms failed in 30d tests. However mTSA and TSA are successful. In the multimodal function ( Layeb03…Layeb18), LSHADE-SPACMA  and mTSA perform well in these functions. In the functions with noise ( Layeb19 and Layeb20), mTSA and TSA are successful in these tests.  It is clear that the proposed test functions are hard; even the mean errors are small, the difference between the solution given by the algorithm and the optimum solution is very high.  For example, in the case of the function Layeb12, the two algorithms LSHADE and LSHADE-SPACMA have mean error lower than 7, however the obtained solutions are far from the global optimum solution, which is [2,2,…2]. So, to illustrate fairly  the performance of an algorithm, it is preferable to incorporate the error between the solution and the global solution if it is known and unique.  So, we propose a new measure called $Mean\ Total\ Error$ (MTE) to evaluate the performance of an algorithm based on both the error function value and the error between the global optimum and the solution found by a given algorithm. This measure is given by the following equation:

$$MTE = \frac{1}{runs}\sum_{i=1}^{runs}(Falgoritm_i - Fglobal) + \|Solutionalgoritm_i - globaloptimum\| \ldots\ldots eq3$$

For the Best Total Error BTE, this equation becomes:

$$BTE = (Falgoritm_{best} - Fglobal) + \|(Solutionalgoritm_{best} - globaloptimum)\| \ldots eq4$$

Where : $\|.\|$ is the Euclidian norm, $Falgoritm_i$ is the value of the algorithm at run $i$. $Solutionalgoritm$ is the solution found by an algorithm.

For the test functions that have multiple global solutions, it's recommended to use a shifted solution in order to get one optimal solution.

By using the new error measure, the performances of the algorithms are clearer compared to the standard mean error. For example, table 3 displays the mean total error of the algorithms for the function Layeb12. It is obviously clear that the solutions given by the algorithms mTSA and TSA are the closest to the exact solution. However, the solutions given by the reaming algorithms are less accurate. The same remark for other test functions.

Table 1: Statistical results of the algorithms for 30d test functions

| Algorithms | | Function name | Layeb01 | Layeb02 | Layeb03 | Layeb04 | Layeb05 | Layeb06 | Layeb07 | Layeb08 | Layeb09 | Layeb10 | Layeb11 | Layeb12 | Layeb13 | Layeb14 | Layeb15 | Layeb16 | Layeb17 | Layeb18 | Layeb19 | Layeb20 |
|---|---|---|---|---|---|---|---|---|---|---|---|---|---|---|---|---|---|---|---|---|---|---|
| | | global optimum | 0 | 0 | -29 | -229.3249 | -200.329 | 0 | 0 | -200.3249 | 0 | 0 | 0 | -29 | -107.8301 | 0 | 0 | 0 | 0 | -200.3249 | 0 | 0 |
| mTSA | mean | | 0.000E+00 | 0.000E+00 | 2.366E+00 | 8.335E+00 | 1.209E+01 | 5.034E+00 | 1.913E+02 | 3.026E+02 | 9.881E-03 | 5.032E-02 | 5.922E+00 | 3.246E-03 | 2.900E+01 | 1.115E+03 | 5.902E+01 | 2.455E+00 | 4.813E+01 | 1.308E+01 | 1.416E-06 | 5.909E-07 |
| | std | | 0.000E+00 | 0.000E+00 | 2.658E+00 | 3.632E+00 | 8.938E+00 | 1.471E+00 | 6.238E+01 | 6.467E+01 | 1.228E-02 | 5.723E-02 | 3.026E+00 | 1.009E-02 | 0.000E+00 | 1.560E+02 | 1.465E+01 | 1.805E+00 | 1.721E+01 | 1.252E+01 | 7.325E-06 | 9.384E-07 |
| | min | | 0.000E+00 | 0.000E+00 | 0.000E+00 | 2.000E+00 | 7.378E-02 | 2.510E+00 | 0.000E+00 | 1.582E+02 | 2.739E-05 | 7.156E-05 | 2.090E+00 | 0.000E+00 | 2.900E+01 | 6.989E+02 | 2.379E+01 | 2.116E-01 | 2.904E+01 | 2.632E-02 | 1.453E-16 | 9.646E-11 |
| | max | | 0.000E+00 | 0.000E+00 | 8.998E+00 | 1.618E+01 | 3.527E+01 | 8.349E+00 | 2.834E+02 | 4.102E+02 | 6.396E-02 | 2.408E-01 | 1.086E+01 | 5.259E-02 | 2.900E+01 | 1.356E+03 | 8.593E+01 | 5.258E+00 | 8.012E+01 | 5.148E+01 | 4.016E-05 | 4.443E-06 |
| TSA | mean | | 0.000E+00 | 1.354E-13 | 2.997E-01 | 8.446E+00 | 6.969E+01 | 5.667E+00 | 6.259E+01 | 3.936E+02 | 6.409E-03 | 1.958E-03 | 7.583E+00 | 5.629E-06 | 2.976E+01 | 9.968E+02 | 1.188E+02 | 3.611E+00 | 4.167E+01 | 2.019E+01 | 2.742E-07 | 2.782E-07 |
| | std | | 0.000E+00 | 3.575E-13 | 5.954E-01 | 2.617E+00 | 2.142E+01 | 1.868E+00 | 8.001E+01 | 9.186E+01 | 4.262E-03 | 6.376E-03 | 4.318E+00 | 4.608E-05 | 4.156E+00 | 1.213E+02 | 2.067E+01 | 2.179E+00 | 2.067E+01 | 1.820E+01 | 1.100E-06 | 6.622E-07 |
| | min | | 0.000E+00 | 0.000E+00 | 0.000E+00 | 2.000E+00 | 2.285E+01 | 2.369E+00 | 0.000E+00 | 2.323E+02 | 1.984E-04 | 2.338E-08 | 2.090E+00 | 0.000E+00 | 2.900E+01 | 7.652E+02 | 7.511E+01 | 3.970E-03 | 2.902E+01 | 8.358E-01 | 7.414E-19 | 8.454E-13 |
| | max | | 0.000E+00 | 1.823E-12 | 1.998E+00 | 1.415E+01 | 9.914E+01 | 9.939E+00 | 2.414E+02 | 5.383E+02 | 1.831E-02 | 2.773E-02 | 1.530E+01 | 1.689E-04 | 5.176E+01 | 1.272E+03 | 1.515E+02 | 7.378E+00 | 1.062E+02 | 7.298E+01 | 5.898E-06 | 3.496E-06 |
| LSHADE | mean | | 1.000E+30 | 9.000E+28 | 2.885E+01 | 2.954E+01 | 6.222E+00 | 1.361E+01 | 1.831E+03 | 3.274E+02 | 4.029E+00 | 1.940E+02 | 8.723E+00 | 5.269E+00 | 5.295E+02 | 4.593E+02 | 1.907E+02 | 1.100E+01 | 4.626E+01 | 8.456E+01 | 4.124E+00 | 1.908E-02 |
| | std | | 1.431E+14 | 3.051E+28 | 3.454E-01 | 8.209E+00 | 2.189E+00 | 2.946E-01 | 3.076E+01 | 2.296E+01 | 4.496E-01 | 3.296E+01 | 5.531E-01 | 5.981E-01 | 2.535E+02 | 8.683E+01 | 2.694E+02 | 3.006E-01 | 3.850E+00 | 3.991E+00 | 3.206E+00 | 8.808E-03 |
| | min | | 1.000E+30 | 4.274E+00 | 2.798E+01 | 1.610E+01 | 2.683E+00 | 1.281E+01 | 1.765E+03 | 2.841E+02 | 3.216E+00 | 1.270E+02 | 7.375E+00 | 4.042E+00 | 3.596E+02 | 3.004E+02 | 1.382E+02 | 1.041E+01 | 3.775E+01 | 7.720E+01 | 2.671E+00 | 6.170E-03 |
| | max | | 1.000E+30 | 1.000E+30 | 2.899E+01 | 4.741E+01 | 9.886E+00 | 1.413E+01 | 1.907E+03 | 3.782E+02 | 4.991E+00 | 2.612E+02 | 9.580E+00 | 6.253E+00 | 1.817E+03 | 6.288E+02 | 6.297E+01 | 1.182E+01 | 5.354E+01 | 9.183E+01 | 1.381E+01 | 4.759E-02 |
| MadDe | mean | | 1.000E+30 | 4.000E+29 | 2.882E+01 | 3.372E+01 | 5.248E+01 | 1.784E+01 | 2.645E+00 | 4.926E+02 | 8.943E+00 | 1.393E+01 | 1.444E+01 | 1.301E+01 | 2.900E+01 | 1.877E+02 | 1.383E+02 | 1.520E+01 | 3.069E+01 | 1.098E+02 | 1.158E+00 | 1.252E-03 |
| | std | | 1.431E+14 | 4.983E+29 | 3.784E-01 | 1.141E+01 | 9.594E+00 | 5.392E-01 | 1.085E+01 | 3.026E+01 | 1.214E+00 | 5.420E-15 | 1.433E+00 | 1.986E+00 | 0.000E+00 | 3.623E+01 | 2.256E+01 | 6.536E-01 | 3.943E+01 | 5.128E+00 | 9.764E-01 | 7.737E-04 |
| | min | | 1.000E+30 | 2.647E+00 | 2.798E+01 | 1.763E+01 | 2.954E+01 | 1.664E+01 | 0.000E+00 | 4.347E+02 | 6.300E+00 | 1.393E+01 | 1.117E+01 | 8.369E+00 | 2.900E+01 | 1.749E+02 | 7.611E+01 | 1.394E+01 | 3.007E+01 | 9.540E+01 | 1.539E-02 | 7.428E-05 |
| | max | | 1.000E+30 | 1.000E+30 | 2.900E+01 | 6.392E+01 | 6.876E+01 | 1.870E+01 | 5.859E+01 | 5.437E+02 | 1.128E+01 | 1.393E+01 | 1.674E+01 | 1.566E+01 | 2.900E+01 | 1.929E+03 | 1.727E+02 | 1.623E+01 | 3.175E+01 | 1.175E+02 | 4.317E+00 | 3.357E-03 |
| LSHADE-SPACMA | mean | | 1.000E+30 | 2.333E+29 | 2.892E+01 | 3.513E+01 | 8.181E+00 | 1.404E+01 | 1.817E+03 | 2.798E+02 | 3.026E+00 | 7.642E+01 | 7.528E+00 | 5.218E+00 | 3.527E+02 | 3.700E+02 | 9.321E+01 | 1.020E+01 | 2.606E+01 | 7.051E+01 | 5.547E-04 | 2.119E-05 |
| | std | | 1.431E+14 | 4.302E+29 | 2.531E-01 | 7.828E+00 | 3.188E+00 | 3.289E-01 | 2.890E+01 | 3.413E+01 | 5.311E-01 | 2.010E+01 | 5.079E-01 | 7.482E-01 | 8.812E+01 | 6.915E+01 | 3.546E+01 | 3.876E-01 | 5.934E+00 | 4.541E+00 | 5.123E-04 | 2.644E-05 |
| | min | | 1.000E+30 | 6.401E-01 | 2.799E+01 | 1.577E+01 | 2.771E+00 | 1.332E+01 | 1.772E+03 | 2.165E+02 | 1.812E+00 | 3.477E+01 | 6.554E+00 | 3.770E+00 | 2.345E+02 | 2.688E+02 | 7.189E+01 | 9.386E+00 | 1.500E+01 | 6.048E+01 | 1.209E-05 | 1.752E-06 |
| | max | | 1.000E+30 | 1.000E+30 | 2.899E+01 | 5.211E+01 | 1.476E+01 | 1.468E+01 | 1.872E+03 | 3.412E+02 | 4.008E+00 | 1.196E+02 | 8.374E+00 | 6.619E+00 | 5.770E+02 | 5.318E+02 | 2.282E+02 | 1.104E+01 | 4.679E+01 | 7.921E+01 | 1.851E-03 | 1.199E-04 |

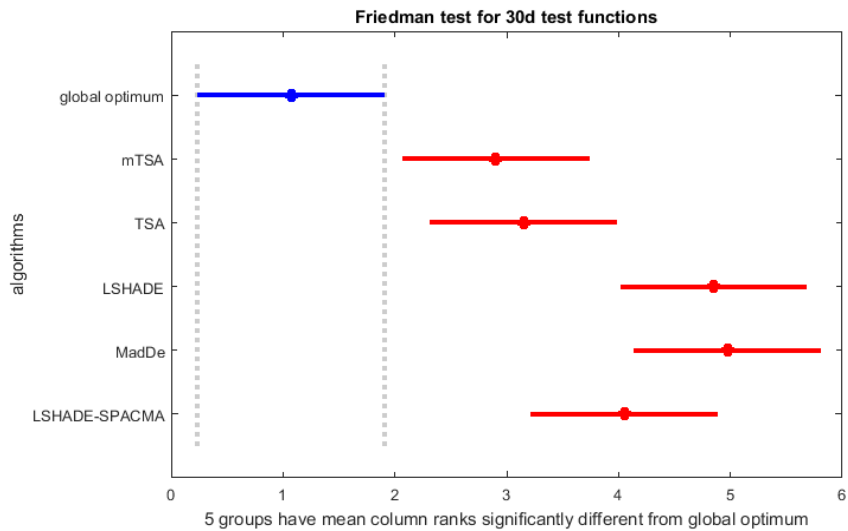

**Table 2:** Statistical results of the algorithms for 10d test functions

| Algorithms | Function name global optimum | | Layeb01 0 | Layeb02 0 | Layeb03 -9 | Layeb04 -71.1698 | Layeb05 -62.169798 | Layeb06 0 | Layeb07 0 | Layeb08 -62.1698 | Layeb09 0 | Layeb10 0 | Layeb11 -9 | Layeb12 -33.464536 | Layeb13 0 | Layeb14 0 | Layeb15 0 | Layeb16 0 | Layeb17 0 | Layeb18 -62.169798 | Layeb19 0 | Layeb20 0 |
|---|---|---|---|---|---|---|---|---|---|---|---|---|---|---|---|---|---|---|---|---|---|---|
| mTSA | mean | | **0.000E+00** | **0.000E+00** | 6.997E-01 | 6.293E+00 | 1.555E+00 | **1.628E+00** | 1.851E+02 | 6.070E+01 | 6.040E-03 | 2.751E-01 | 1.852E+00 | **1.473E-01** | 2.742E+01 | 2.900E+02 | 1.768E+01 | **2.813E+00** | 1.304E+01 | 7.927E+00 | 7.612E-06 | 5.227E-06 |
| | std | | 0.000E+00 | 0.000E+00 | 9.876E-01 | 6.796E+00 | 3.128E+00 | 6.838E-01 | 8.575E+01 | 1.570E+01 | 1.027E+01 | 4.581E-01 | 1.086E+00 | 2.578E-01 | 5.157E+01 | 5.142E+01 | 9.147E+00 | 1.175E+00 | 5.364E+00 | 5.417E+00 | 3.841E-05 | 9.799E-06 |
| | min | | 0.000E+00 | 0.000E+00 | 0.000E+00 | 0.000E+00 | 4.313E-01 | 3.620E-01 | 0.000E+00 | 1.559E+01 | 1.416E-05 | 7.502E-04 | 5.214E-02 | 3.892E-09 | 9.000E+00 | 1.590E+02 | 7.134E+00 | 1.141E+00 | 9.049E+00 | 2.457E-01 | 3.864E-16 | 5.073E-10 |
| | max | | 0.000E+00 | 0.000E+00 | 3.999E+00 | 2.196E+01 | 8.093E+00 | 2.994E+00 | 3.782E+02 | 7.981E+01 | 4.268E-02 | 2.200E+00 | 3.436E+00 | 9.118E-01 | 2.264E+02 | 3.774E+02 | 4.203E+01 | 5.574E+00 | 3.156E+01 | 2.165E+01 | 2.106E-04 | 3.586E-05 |
| TSA | mean | | **0.000E+00** | 1.482E-02 | **6.666E-02** | 5.995E+00 | 1.490E+01 | 2.054E+00 | **6.936E-06** | 7.377E+01 | **3.263E-03** | **4.022E-02** | 3.124E+00 | **3.362E-02** | 3.398E-02 | 3.074E+00 | 3.485E+01 | 3.948E+00 | 1.444E+01 | 9.245E+00 | 5.620E-05 | 2.133E-06 |
| | std | | 0.000E+00 | 8.114E-02 | 2.537E-01 | 8.279E+00 | 7.786E+00 | 1.256E+00 | 2.922E-05 | 3.177E+01 | 5.503E-03 | 1.591E-01 | 1.464E+00 | 1.075E-02 | 3.580E-02 | 6.000E+01 | 9.319E+00 | 1.571E+00 | 2.008E+01 | 7.540E+00 | 3.061E-04 | 3.952E-06 |
| | min | | 0.000E+00 | 0.000E+00 | 0.000E+00 | 1.479E+00 | 3.196E-02 | 0.000E+00 | 1.498E+01 | 9.696E-07 | 9.235E-07 | 6.486E-01 | 4.346E-09 | 2.379E-04 | 1.919E+02 | 1.382E+00 | 9.000E+00 | 3.374E-01 | 2.220E-18 | 2.486E-11 | | |
| | max | | 0.000E+00 | 4.444E-01 | 9.999E-01 | 2.445E+01 | 3.237E+01 | 4.683E+00 | 1.580E-04 | 1.470E+02 | 2.532E-02 | 8.549E-01 | 5.230E+00 | 4.134E-01 | 1.705E-01 | 4.116E+02 | 5.511E+01 | 6.567E+00 | 1.148E+02 | 2.168E+01 | 1.677E-03 | 1.504E-05 |
| LSHADE | mean | | 7.000E+29 | 4.415E+00 | 4.220E-01 | **4.221E-01** | **1.236E-02** | 2.264E+01 | 4.765E+02 | 4.323E+01 | 2.127E-01 | 3.609E+00 | 1.106E+00 | 4.708E-01 | 2.914E+01 | 4.552E+01 | 1.534E+01 | 4.491E+00 | 9.008E+00 | 1.295E+01 | 3.913E-03 | 3.312E-05 |
| | std | | 4.661E+29 | 1.128E-01 | 8.122E-01 | 8.085E-01 | 1.428E+01 | 1.097E+01 | 9.027E+02 | 2.408E+00 | 2.638E-01 | 1.173E-01 | 9.355E+01 | 6.200E+01 | 5.670E+00 | 2.955E-01 | 2.071E+02 | 2.108E+00 | 8.146E-03 | 2.789E-05 | | |
| | min | | 0.000E+00 | 4.000E+00 | 0.000E+00 | -8.392E-10 | 1.289E-03 | 7.311E-01 | 4.124E-09 | 1.644E+01 | 6.233E-08 | 9.034E-01 | 3.962E-01 | 1.574E-01 | 3.673E+01 | 7.536E+00 | 5.928E+00 | 4.031E+00 | 9.000E+00 | 9.402E-01 | 1.925E-05 | 3.743E-06 |
| | max | | 1.000E+30 | 4.444E+00 | 2.997E+00 | 2.000E+00 | 3.491E-02 | 3.430E+01 | 5.156E+02 | 6.426E+01 | 4.113E-01 | 9.647E+00 | 1.726E+00 | 6.574E-01 | 4.849E+02 | 3.475E+02 | 3.208E+01 | 5.162E+00 | 9.069E+00 | 1.666E+01 | 4.289E-02 | 1.004E-04 |
| MadDe | mean | | **0.000E+00** | 3.407E-01 | 4.397E+00 | 1.675E+00 | 1.302E+02 | 2.399E+00 | 3.105E+02 | 3.768E+01 | 6.868E-02 | 4.324E+00 | **6.071E-01** | 5.378E-01 | 9.000E+00 | **2.829E+02** | **2.817E+00** | 2.993E+00 | **4.405E+00** | **7.368E+00** | 6.305E-03 | 4.185E-05 |
| | std | | 0.000E+00 | 4.157E-01 | 1.215E+00 | 1.442E+00 | 1.562E+02 | 7.380E-01 | 8.410E+01 | 1.748E+01 | 6.962E-02 | 0.000E+00 | 4.598E-01 | 1.599E-01 | 0.000E+00 | 5.118E+01 | 2.870E+00 | 7.422E-01 | 3.519E+00 | 3.256E+00 | 7.612E-03 | 3.455E-05 |
| | min | | 0.000E+00 | 0.000E+00 | 1.998E-01 | 0.000E+00 | 9.102E-06 | 1.030E+00 | 1.378E+02 | 1.490E+00 | 5.912E-04 | 4.324E+00 | 7.218E-04 | 1.439E-01 | 9.000E+00 | 1.464E+02 | 4.124E-03 | 1.394E+00 | 6.327E-07 | 1.893E+00 | 1.994E-04 | 2.176E-06 |
| | max | | 0.000E+00 | 1.333E+00 | 6.949E+00 | 5.562E+00 | 6.599E-02 | 3.802E+00 | 4.825E+02 | 6.408E+01 | 2.175E-01 | 4.324E+00 | 1.146E+00 | 8.463E-01 | 9.000E+00 | 7.777E+02 | 6.818E+00 | 4.237E+00 | 9.166E+00 | 1.397E+01 | 3.369E-02 | 1.462E-04 |
| LSHADE-SPACMA | mean | | **0.000E+00** | 2.978E+00 | 3.556E-01 | 2.410E+01 | 1.435E+00 | 2.421E+00 | 4.690E+02 | **2.200E+01** | 4.927E-02 | 5.918E-01 | 9.434E-01 | 5.364E-01 | **6.614E-17** | 7.193E+00 | 9.144E+00 | 4.042E+00 | 8.474E+00 | 9.079E+00 | 7.489E-09 | 5.301E-07 |
| | std | | 0.000E+00 | 1.978E+00 | 6.073E-01 | 2.504E+00 | 1.194E+00 | 4.542E-01 | 1.594E+01 | 1.198E+00 | 3.277E-02 | 5.155E-01 | 3.280E-01 | 1.084E-01 | 2.758E-16 | 5.849E+00 | 7.035E+00 | 3.441E-01 | 1.448E+00 | 1.724E+01 | 3.251E-08 | 8.083E-07 |
| | min | | 0.000E+00 | 0.000E+00 | 0.000E+00 | 1.405E+01 | 7.648E-04 | 1.534E+00 | 4.332E+02 | 1.473E+01 | 3.686E-03 | 7.964E+00 | 1.950E-01 | 2.817E-01 | 0.000E+00 | 1.144E-01 | 1.803E+00 | 3.279E+00 | 4.000E+00 | 5.138E+00 | 2.029E-15 | 8.325E-09 |
| | max | | 0.000E+00 | 4.444E+00 | 2.040E+00 | 2.500E+01 | 4.102E-02 | 3.147E+00 | 4.976E+02 | 4.919E+01 | 1.268E-01 | 1.711E+00 | 1.381E+00 | 7.899E-01 | 1.488E-15 | 2.522E+01 | 2.916E+01 | 4.589E+00 | 9.931E+00 | 1.206E+01 | 1.790E-07 | 3.984E-06 |

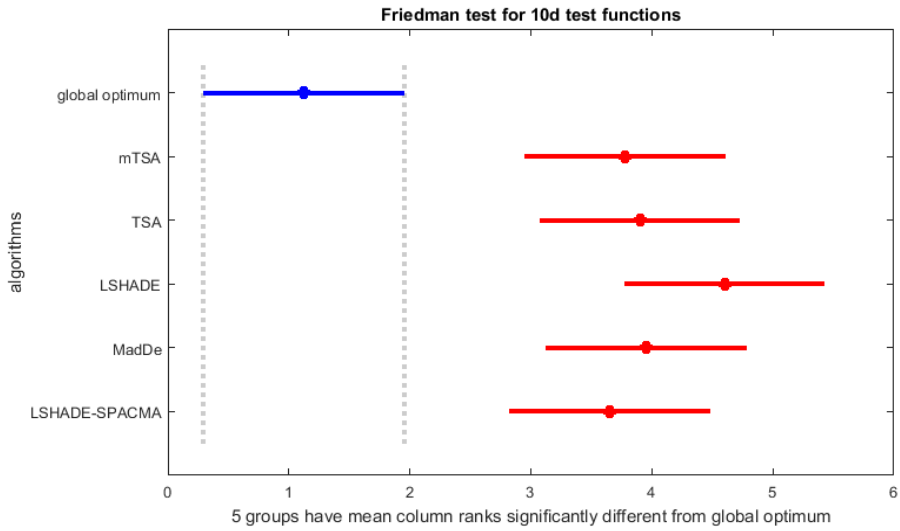

**Figure 3.** Freidman test comparison for 10d test functions

**Table 3.** Mean error vs mean total error for the Layeb12 function.

|  | **Mean error** | **Mean total error** |
|---|---|---|
| **mTSA** | 0.003 | 0.232 |
| **TSA** | 0.000 | 0.023 |
| **LSHADE** | 5.270 | 29.999 |
| **MaDde** | 13.010 | 31.112 |
| **LSHADE-SPACMA** | 5.218 | 19.596 |

### 4. Conclusion

In this work, we have proposed some new functions divided into unimodal functions, multimodal functions, and functions with noise. The experimental study has proven the hardness of the proposed function. The proposed test functions also validate the "no free lunch" theorem that stipulates that no algorithm is efficient to deal with all the optimization problems. We have seen that top rank algorithms in CEC competitions are not so successful in dealing with the proposed test functions.

# Annex

## modified Tangent Search algorithms (mTSA)

```
clc;
clear;
(rand('state', sum(100*clock))); %#ok<RAND>
%layebfnunresults=zeros(20,34); % save results
%addpath('./layeb_benchmark');
Runs=1; % nbr of runs
 optimums30d=[0 0   -29 -229.324    -200.329   0   0  -200.324   0   0  -29 -107.830
0   0   0   0   0   -200.324    0   0]; %30d
%optimum10d= [ 0    0   -9 -71.169    -62.169    0   0  -62.169    0   0  -9 -33.464
0   0   0   0   0   -62.169    0   0]; %10d

for nfun=1:20
    %---------------test function parameters------------------
    func = callFunction(nfun);  %get the function struct for [1...50] functions
    fun =str2func( func.name);   % function to be optimized
    %fun =str2func(strcat(func.name ,'_mex'));    % mex function for speed
    dim =func.dim; %dimension of the problem
    lb =  func.lowerlimit; %lower limit of the problem
    ub = func.upperlimit; %upper limit of the problem

    fprintf('\n-------------------------------------------------------\n')
    fprintf('Function = %s, Dimension size = %d\n', func.name, dim)
    for nbex=1:Runs %
        (rand('state', sum(100*clock))); %#ok<RAND>

     %----------------STO paramaters
       nbr_agent=40;       %population size
       FES=0;
       MAX_FES=10000;   % maximum number of function evaluation

     %-------------------------------
       pop=struct('fitness',{},'position',{}); % initialization of population
       %fitness_pop=zeros(1,nbr_agent);

       for i=1:nbr_agent
           X=lb+(ub-lb).*rand(1,dim);    % generate a new solution X
           % fitness=feval(fun,X',nfun);
           fitness=fun(X);
           pop(i).position=X;
           pop(i).fitness=fitness;
           % fitness_pop(i)=fitness;
           FES=FES+1;

           if i==1
               Bestagent=pop(1);
           end

           if pop(i).fitness< Bestagent.fitness
               Bestagent=pop(i);
               Bestagent.fitness=pop(i).fitness;
           end

       end

    %%---------------- begin  iterations
       while FES<=MAX_FES    % iteration process
           if  FES>MAX_FES
               break;
           end

           for j=1:nbr_agent
               optx=Bestagent.position;
               X=pop(j).position;

      %--------------------exploration phase---------------

              for jk=1:dim

                  teta=rand*pi/2.5;
                  id=randi(dim);
```

```matlab
                        if rand<=1.5/dim   || jk==id %1.5/dim best
                            if isequal (optx,X)
                                step=0.1*sign(rand-0.5)/log(1+FES);
                                X(jk)=X(jk)+step*(tan(teta));
                            else

                                step=0.5*sign(rand-0.5)*norm(optx-X);

                                if rand  <=0.3
                                    X(jk)=X(jk)+tan(rand*(pi));% large tangent flight
                                else
                                    X(jk)=X(jk)+step*(tan(teta)); % small tangent flight

                                end

                            end

                        end
                    end
            %------- bounds checking
                Xnew=X;
                Xnew(Xnew>ub)=rand*(ub - lb) + lb;
                Xnew(Xnew<lb)=rand*(ub - lb) + lb;
            %----------evaluation and best solution update

                fitness= fun(Xnew);
                % fitness= feval(fun,Xnew',nfun);
                if   fitness <pop(j).fitness
                   pop(j).position   =Xnew ;
                   pop(j).fitness=fitness;
                  if   fitness< Bestagent.fitness
                   Bestagent.position=Xnew;
                   Bestagent.fitness=fitness ;
                  end
                end
                FES= FES+1;
                if  FES>MAX_FES
                    break;
                end

    %--------------------end exploration phase--------------------

     %-------------------- intensification search--------------------

                if    (rand<0.7 && FES >=0.5*MAX_FES) ||  (rand<0.05)  % 0.2 for test 3

                    X=pop(j).position;
                    B=X;
                    teta=rand*pi/2.5;
                    step = 1*sign(rand-0.5)*norm(optx)*log(1+10*dim/FES);
                    if isequal(optx, X)
                        X=optx+step.*(tan(teta)).*(rand*optx-X);
                    else
                        if rand <=0.7
                            X=optx+step.*(tan(teta)).*(optx-X);
                        else
                            sign1 = -1+(1-(-1)).*rand();
                            ro = 15*sign1*1/log(1+FES);
                            X = X + ro.*(optx-rand*(optx-X)) ;
                        end

                    end

                    pm=randperm(dim);
                    if dim<=4
                        pm2=round(0.4*dim); %0.1
                    else
                        pm2=round(0.2*dim); %0.2
                    end
                    if rand <=1
                        X(pm(1:pm2))=B(pm(1:pm2));
                    else
                        X(pm(1:pm2))=optx(pm(1:pm2));
                    end
```

```matlab
                %----bounds checking
                Xnew=X;
                Xnew(Xnew>ub)=rand*(ub - lb) + lb;
                Xnew(Xnew<lb)=rand*(ub - lb) + lb;
                %---------evaluation and best solution update
                fitness= fun(Xnew);
                % fitness= feval(fun,Xnew',nfun);
                if   fitness <pop(j).fitness
                    pop(j).position   =Xnew ;
                    pop(j).fitness=fitness;
                   if   fitness< Bestagent.fitness
                    Bestagent.position=Xnew;
                    Bestagent.fitness=fitness ;
                   end
                end
                FES= FES+1;
                if  FES>MAX_FES
                    break;
                end
                end
    %-------------------- end intensification phase------------------

    %-----------------escape local minimma----------------------------
                if rand <0.01 %default0.01
                    X=pop(j).position;
                    B=X;
                    teta=rand*pi; %randn
                    X =   X + tan(teta).*(ub-lb); % generate a random solution by tangent flight
                    Xnew=X;
                %------- bounds checking
                    if rand <=0.8
                       Xnew(Xnew>ub)=rand*(ub - lb) + lb;
                       Xnew(Xnew<lb)=rand*(ub - lb) + lb;
                    else
                       Xnew(Xnew>ub)=((rand*(ub - lb) + lb)+B(Xnew>ub))/2;
                       Xnew(Xnew<lb)=(rand*(ub - lb) + lb+B(Xnew<lb))/2;
                    end

                %---------evaluation and best solution update
                fitness= fun(Xnew);
                % fitness= feval(fun,Xnew',nfun);
                if   fitness <pop(j).fitness
                    pop(j).position   =Xnew ;
                    pop(j).fitness=fitness;
                   if   fitness< Bestagent.fitness
                    Bestagent.position=Xnew;
                    Bestagent.fitness=fitness ;
                   end
                end
                FES= FES+1;
                if  FES>MAX_FES
                    break;
                end

                end

            end%j

        end%while

        %fprintf(' Valopt = %d, fun=  %s, nbex=  %d  \n ',Bestagent.fitness,func.name,nbex)
        fprintf(' error = %d, fun=  %s, nbex=  %d  \n ',Bestagent.fitness-optimums30d(nfun),func.name,nbex)

        %layebfnunresults(nfun,nbex)=Besttrochoid.fitness-optimum30d(nfun);
        %layebfnunresults(nfun,nbex)=Bestagent.fitness;
        % MTE(nfun,nbex)=Besttrochoid.fitness-optimum(nfun)+ norm(globalbest-optx);
    end
end
% for i=20:20
```

```
%       layebfnunresults(i,31)=(mean(layebfnunresults(i,1:30)));
%       layebfnunresults(i,32)=std(layebfnunresults(i,1:30));
%       layebfnunresults(i,33)=min(layebfnunresults(i,1:30));
%       layebfnunresults(i,34)=max(layebfnunresults(i,1:30));
% end
%
% xlswrite('results.xlsx',layebfnunresults,1)
```